%% file: Detailed3Dreconstruction.tex
\documentclass[10pt,journal,compsoc]{IEEEtran}
\ifCLASSOPTIONcompsoc
  \usepackage[nocompress]{cite}
\else
  \usepackage{cite}
\fi
\ifCLASSINFOpdf
\else
\fi

\usepackage{times}
\usepackage{epsfig}
\usepackage{graphicx}
\usepackage{amsmath}
\usepackage{amssymb}
\usepackage{caption,subcaption,sidecap}
\usepackage{textcomp,bm,color,mathrsfs}
\usepackage{mwe}
\usepackage[boxed]{algorithm2e}
\usepackage[english]{babel}

\newcommand{\tabincell}[2]{\begin{tabular}{@{}#1@{}}#2\end{tabular}}

\usepackage[pagebackref=false,breaklinks=true,letterpaper=true,colorlinks,bookmarks=false]{hyperref}

\begin{document}
%
\title{Detailed Surface Geometry and Albedo Recovery from RGB-D Video under Natural Illumination}
%
%
%
%

\author{Xinxin~Zuo,
        Sen~Wang,
        Jiangbin~Zheng,
        Zhigeng~Pan,
        and~Ruigang~Yang,~\IEEEmembership{Senior~Member,~IEEE}
\IEEEcompsocitemizethanks{\IEEEcompsocthanksitem X. Zuo is with the Northwestern Polytechnical University, Xi'an, China; University of Kentucky, Lexington, KY, USA. E-mail: xinxin.zuo@uky.edu 
\IEEEcompsocthanksitem S. Wang and J. Zheng are with Northwestern Polytechnical University, Xi'an, China. E-mail: wangsen1312@gmail.com, zhengjb@nwpu.edu.cn
\IEEEcompsocthanksitem Z. Pan is with the DMI Research Center, Hangzhou Normal University, Hangzhou, Zhejiang, China. E-mail: zgpan@hznu.edu.cn
\IEEEcompsocthanksitem R. Yang is with University of Kentucky, Lexington, KY, USA; the Baidu Research, Beijing, China; National Engineering Laboratory of Deep Learning Technology and Application, China. E-mail: ryang@cs.uky.edu
}
\thanks{Manuscript received...}}

%

\markboth{IEEE TRANSACTIONS ON PATTERN Analysis and Machine Intelligence,~Vol.~XX, No.~XX, Month~Year}%
{Shell \MakeLowercase{\textit{et al.}}: Bare Demo of IEEEtran.cls for IEEE Journals}
%



\IEEEtitleabstractindextext{%
\begin{abstract}
This paper presents a novel approach for depth map enhancement from an RGB-D video sequence. The basic idea is to exploit the photometric information in the color sequence to resolve the inherent ambiguity of shape from shading problem. Instead of making any assumption about surface albedo or controlled object motion and lighting, we use the lighting variations introduced by casual object movement. We are effectively calculating photometric stereo from a moving object under natural illuminations. One of the key technical challenges is to establish correspondences over the entire image set. We therefore develop a lighting insensitive robust pixel matching technique that out-performs optical flow method in presence of lighting variations. An adaptive reference frame selection procedure is introduced to get more robust to imperfect lambertian reflections. In addition we present an expectation-maximization framework to recover the surface normal and albedo simultaneously, without any regularization term. We have validated our method on both synthetic and real datasets to show its superior performance on both surface details recovery and intrinsic decomposition.
\end{abstract}

}

\maketitle

\IEEEdisplaynontitleabstractindextext

%
\IEEEpeerreviewmaketitle

\input{intro}
\input{related}
\input{Preliminary}
\input{pipeline}

\input{algorithm}
\input{exp}
\input{conclusion}


%



\vspace{-5pt}
\ifCLASSOPTIONcompsoc
  \section*{Acknowledgments}
\else
  \section*{Acknowledgment}
\fi
This work is partially supported by the USDA grant (2018-67021-27416), US NFS (IIP-1543172), Chinese National Key R\&D project (2017YFB1002803), NSFC (No. 61972321), Innovation Chain of Shaanxi Province Industrial Area (2017ZDXM-GY-094), NSERC Discovery Grant (No. RGPIN-2019-04575), the University of Alberta-Huawei Joint Innovation collaboration grant (No. 201902). Jiangbin Zheng and Ruigang Yang are the co-corresponding authors for this paper.

\ifCLASSOPTIONcaptionsoff
  \newpage
\fi



%
\bibliographystyle{IEEEtrans}
\bibliography{detiled3D}

%

\begin{IEEEbiography}
	[{\includegraphics[width=1in,height=1.25in,clip,keepaspectratio]{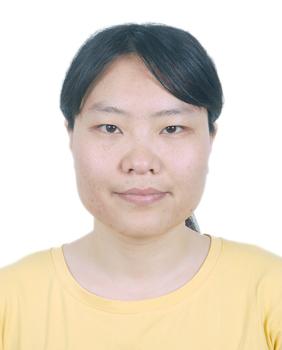}}]
	{Xinxin Zuo} received the B.E. and M.E. degrees from Northwestern Polytechnical University, Xi’an, China, in 2011 and 2014, respectively. She is currently pursuing the Ph.D. degree with the University of Kentucky. Her research interests include computer vision and graphics, especially on 3D reconstruction and human modeling.
\end{IEEEbiography}

\begin{IEEEbiography}
	[{\includegraphics[width=1in,height=1.25in,clip,keepaspectratio]{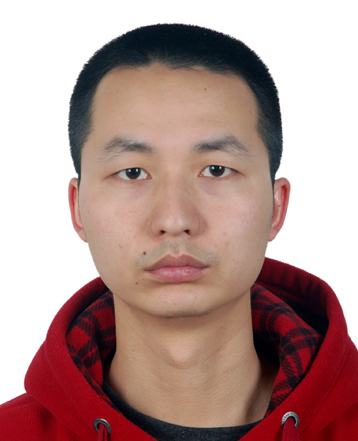}}]
	{Sen Wang} received the B.E. degree from Northwestern Polytechnical University, Xi’an, China, in 2011, where he is currently pursuing the Ph.D. degree. From 2015 to 2016, he was a Visiting Ph.D. Student at the University of Kentucky. His research interests include robotics and computer vision. 
\end{IEEEbiography}

\begin{IEEEbiography}
	[{\includegraphics[width=1in,height=1.25in,clip,keepaspectratio]{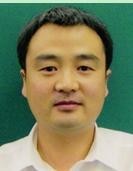}}]
	{Jiangbin Zheng} received the Ph.D. degree from Northwestern Polytecnical University, in 2002, where he is a Full Professor and Dean with School of Software. His research interests include computer graphics, computer vision and multimedia. He has published over 100 papers in the above related research area.
\end{IEEEbiography}

\begin{IEEEbiography}
	[{\includegraphics[width=1in,height=1.25in,clip,keepaspectratio]{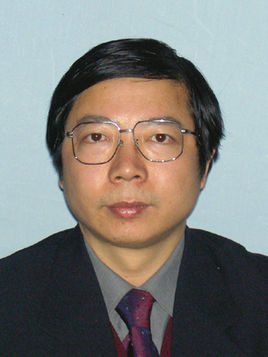}}]
	{Zhigeng Pan} obtained his Bachelor and Master degrees from the Computer Science Department at Nanjing University in 1987 and 1990, respectively, and his PhD degree in 1993 from Zhejiang University. Currently he is the director of Digital Media and HCI Research Center at Hangzhou Normal University. His research interests include HCI, virtual reality, and digital entertainment. He has published over 100 papers in the related research area and received several awards such as second-level National award on Achievements in Science and Technology.
\end{IEEEbiography}


\begin{IEEEbiography}
	[{\includegraphics[width=1in,height=1.25in,clip,keepaspectratio]{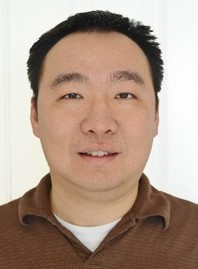}}]{Ruigang Yang (SM'13)}
	received the M.S. degree from Columbia University and the Ph.D. degree from the University of North Carolina at Chapel Hill. He is currently a Full Professor in computer science with the University of Kentucky. He has published over 100 papers, which, according to Google Scholar, has received close to 10000 citations with an h-index of 48 (as of 2017). His research interests include computer graphics and computer vision, in particular in 3D reconstruction and 3D data analysis. He has received a number of awards, including the US NSF CareerAward, in 2004 and the Dean’s Research Award from the University of Kentucky, in 2013. He is currently an Associate Editor of IEEE TPAMI.
\end{IEEEbiography}




\end{document}

%% file: intro.tex
\section{Introduction}
\IEEEPARstart{T}HE availability of affordable depth sensors has sparked a revolution in many areas of computer vision, such as human computer interactions, robotics, and video analysis. Among these 3D modeling has probably received the most benefits from this advancement of sensors. Nevertheless, the current generation of depth sensors, such as the Microsoft Kinect, Asus Xtion and Intel Real-Sense, still suffers from limited resolution and accuracy. As a result, fine-scale structural details of an object cannot get recovered from those low quality depth sensors.

On the other hand, a high resolution color camera is readily available such as the current cellphones. Therefore lots of depth denoising and up-sampling approaches~\cite{Park14,Ferstl13,Yang14,Bohme10,Min12} have been proposed taking a corresponding high quality color image as guidance to enhance the depth map. Among them, some researchers have exploited the consistency or edge correlation between depth and color images to get a noise-free depth image with clear surface edges. However, the detailed structural information is still unrevealed after the denoising operation. There are other methods that take advantage of the shading information contained in color images so as to recover geometric surface details. Most of these shading based methods implement shading refinement on a single RGB-D frame on the basis of Shape-from-Shading (SfS) techniques~\cite{Yu13,Or-el15,Wu14}. However, the inherent ambiguity for SfS still exists. It is an ill-posed problem to recover the surface normal from the measured intensity of a single pixel. Therefore smoothness terms or some regularization terms are usually incorporated.

Another problem of applying the shape from shading technique is that an albedo image is needed to predict the appearance and separate the texture layer from the shading image, which is also unknown. Typically a constant albedo assumption is made, which is not applicable in practice. Therefore, in order to handle varying albedos, there have been lots of researches on this topic as so called intrinsic image decomposition~\cite{Barrow78}. They focus on analyzing the different characteristics of the appearance and shading layers with lots of priors or regularization constraints proposed. Similarly, previous shading-based refinement methods~\cite{Or-el15,Wu14} often approached this chicken-and-egg problem by assuming prior assumptions or by enforcing particular albedo regularizers. However, these regularizers are heuristic and may not work all the time. In this paper we propose to utilize an RGB-D sequence to uniquely recover the normal maps and albedo images altogether without relying on any regularization.


We capture the RGB-D sequence with a Kinect V2 depth sensor attached with a relatively high quality color camera. During the acquisition process, we can rotate the object casually in front of the cameras with the depth and color cameras being static.
In this way, the illumination changes in the image sequence induced by the object's movement provides us the valuable shading correlation along the sequence, which is critical to resolve the surface normal and albedo without any ambiguity.
It resembles the photometric stereo. But instead of controlling the light when imaging a static object, we are allowed to move the object under general natural lighting. 
This kind of cue has been exploited in multi-view photometric stereo~\cite{Esteban08} and shape from video~\cite{Lakdawalla07,Simakov03,zhang03}. However, they have the environmental lighting constrained to be calibrated directional light and the object is experiencing turntable motion or the motion is assumed be calibrated beforehand. On the contrary our approach works under natural lighting with the object experiencing arbitrary motion using a single RGBD sensor, which makes our method more widely used in everyday environment.

Given the captured RGB-D sequence, first we try to align the RGB-D sequence and find the correspondences among the images using a novel robust matching technique. Then the environmental lighting is estimated using the intensity ratios of the aligned sequence, which effectively factors out the impact of varying albedo. Finally, we formulate an Expectation-Maximization based framework in which the surface normal and its albedo map can be calculated robustly, in the presence of some non-Lambertian reflection or cast shadow. A detailed surface mesh is obtained after integration of the initial depth map with the estimated normal map.

The main contribution is that we utilize the dynamic photometric information along the sequence to recover the surface details beyond the resolution of current depth sensors.
Compared to previous depth enhancement schemes that use the color information, our method, to the best of our knowledge, is the least restrictive. It allows arbitrary surface albedo, does not require controlled or calibrated lighting or turntable capture. To achieve these, we make two technical contributions. The first is a novel image registration scheme that is robust to lighting variations and the second is an EM optimization scheme to produce per-pixel normal and albedo map under general lighting.

This paper extends our previous work~\cite{Zuo17}. Specifically, we have proposed an adaptive reference frame selection approach before the image registration and normal computation procedures. Instead of always taking the first image as the reference frame~\cite{Zuo17}, we would like to find out the optimal regions along the sequence that could be marked as reference region. Practically, the sampling rate of the same region on the surface changes as we take images under various rotation of the object. Taking this into consideration, we want to select the reference regions adaptively where the surface details have been better revealed in the image set. Besides, some artifacts such as cast shadow or specular reflections contained in the reference region will prevent the image registration procedure from finding real correspondences along the sequence. Since these artifacts are often view dependent, we can get rid of those artifacts with the adaptive selection approach by selecting reference regions which are free of those artifacts. In this way, we can be more robust to non-lambertian reflectance. More comparison results are also demonstrated in the experiments with complete evaluations of the key components of the techniques. 

%% file: related.tex
\section{Related Work}
In this section, we will review the previous works in two related topics: surface geometry recovery or depth enhancement with shading information, and intrinsic image decomposition.

\subsection{Shape from shading and photometric stereo for surface or depth enhancement} \label{section21}

The Shape-from-shading (SfS) problem has long been studied since the pioneering work by Horn~\cite{Horn70}. It aims to estimate surface normal (and then indirectly surface shape) from a single image. There are various regularization terms or prior assumptions~\cite{Barron11,Barron15} that have been enforced to deal with the inherently ill-posed problem. For example, as one of the state-of-the-art approaches on single image SfS estimation, Barron~\cite{Barron15} obtain strong priors by training from reflectance and shading exemplers and try to recover the reflectance, shape and illumination in a unified framework.

Besides of the SfS recovery on a single image, some methods~\cite{Han13,Yu13} have shown that SfS can be used to refine the noisy depth map captured from RGB-D cameras as well. The inherent ambiguity of SfS is not resolved exactly, but with the initial depth close to the real surface, Wu~\cite{Wu14} and Roy~\cite{Or-el15} have achieved good performance in recovering surface details. More recently, Haefner~\cite{Haefner18} has combined heterogeneous depth and color data to jointly solve the ill-posed depth super-resolution and shape from shading problems. Varying albedo poses another challenge as it needs to be factored out before lighting estimation and shading refinement. Some~\cite{Han13} assumes uniform or piecewise constant albedo. Yu~\cite{Yu13} deals with this by clustering a fixed set of discrete albedos before optimizing geometry. A better, yet more complex strategy, is to simultaneously optimize for unknown albedos and refine geometry~\cite{Kim15}. There are also previous works that adopt the shading constraints to improve the coarse 3D shape reconstructed using multi-view stereo~\cite{Wu11}. More recently, Maier~\cite{intrinsic3d} has proposed to optimize the geometry encoded in a signed distance function, textures and SfS refinement in an unified framework under estimation of spatially-varying spherical harmonics which has achieved the state-of-the-art results on reconstructed scene geometry.

The major difference between our method and these prior works is that we fully exploit lighting effects contained in a video sequence to uniquely and simultaneously determine the surface albedo and normal in a pixel-wise manner. Therefore we are able to deal with arbitrary albedo.


Our approach is also related to photometric stereo methods which have been developed to compute the surface normal using multiple images of a scene taken under different or controlled illumination~\cite{Wu10,Wu06dense}. Unlike SfS, photometric stereo is a well-defined problem, which can be incorporated to enhance the raw depth map~\cite{Haque14, Zhang12} captured from the current depth sensor. Wu~\cite{Wu11fusing} and Zhou~\cite{zhou13multi} have also shown its effectiveness under multi-view setup or for images acquired from the internet~\cite{Shi14}. Basri~\cite{Basri2007Photometric} has proposed to solve the photometric stereo problem under general and unknown lighting using matrix decomposition. Besides, some have used the IR images instead of color images for normal estimation~\cite{Ti15}. Recently Chatterjee~\cite{Chatterjee15} exploited the IR image and proposed a factorization method to handle objects with varying albedo. In these approaches, the objects are kept to be static or captured under almost the same viewpoint with different lighting conditions. Different from them we allow the object to rotate arbitrarily under uncontrolled environment.

\subsection{Intrinsic decomposition} 
\label{section22}

There are some works about intrinsic decomposition that focus on separating the albedo map from shading image, to which our work is also related. The problem of intrinsic image decomposition, first introduced by Barrow~\cite{Barrow78}, is to separate reflectance and illumination from a single image. It is again an ill-posed problem since there are two unknowns for every observation. Additional constraints must be adopted to make the problem solvable. The Retinex theory~\cite{Retinex71} is widely used for this purpose. It assumes that shading variations are mostly low frequency while albedo changes are mostly high-frequency. Based on this assumption, many approaches have been proposed. For example, Tappen et al.~\cite{Tappen05} train a classifier from reflectance and shading data sets. Global priors, such as the reflectance sparsity, are developed~\cite{Garces12,Bi15,Zhao12} and they perform clustering or enforce non-local smoothness on image reflectance. These priors or regularization terms are not guaranteed to work well in all cases, especially when the albedo variation is significant (e.g., a textured surface).


More information has been used to reduce the ambiguity. For example, researchers propose to use RGB-D image as input and depth maps or surface normals are taken as additional cues~\cite{Chen13,Jeon14,Barron16}. Lee et al.~\cite{Lee12} employed temporal albedo consistency constraints for RGB-D video. Another solution is to leverage multiple images of the scene taken under varying lighting conditions~\cite{Kong14, Laffont12, Laffont15}. However, without any information about the surface geometry and environmental lighting, the problem is still ill-posed with regularization terms or pairwise constraints needed.

The main difference between our method and those prior works is that we can solve normal and albedo for each pixel independently, \emph{without the need for additional regularization terms}.

In addition to the above conventional optimization or filtering solutions with strong prior assumptions, deep learning based approaches have also been proposed to compute intrinsic image decomposition when granted access to sufficient labeled training data. Since this is beyond the scope of this paper, we refer to a recent paper~\cite{bonneel2017intrinsic} for a comprehensive overview of current methods on single image intrinsic decomposition exploiting the deep learning techniques.




%



%% file: Preliminary.tex
\section{Preliminary Theory}
Before introducing our proposed approach in detail, we demonstrate some basic theory and derivations in this section.

While environmental lighting can be arbitrarily complex, the appearance of a diffuse object can be described by a low dimensional model~\cite{Ramamoorthi01}. Under this assumption, the shading function $s$ for Lambertian reflectance can be modeled as a quadratic function of the surface normal with $A$, $b$, $c$ represented as the lighting parameters.
\begin{equation}
	s(\bm{n}) = \bm{n}^{T} A \bm{n} + b^{T} \bm{n} + c
\end{equation}

Generally the captured image is generated by multiplying the shading function with surface albedo $\rho(p)$
\begin{equation}
	I(p) = \rho(p) s(\bm{n}_p)
\end{equation}

Given a single image as observation, for each pixel we have three equations with five unknowns to be estimated, it may not be feasible to recover the surface normal and albedo faithfully even though we could suppose the lighting parameters have been determined beforehand. Photometric stereo, with more lighting variations, is a typical solution to resolve the ambiguity. Mathematically, the surface normal and its albedo can be computed by minimizing the following objective function that is formulated for each pixel independently under various lighting conditions. $A_k$, $b_k$, $c_k$ is one set of lighting parameters and $k$ can goes from $1$ to $N$ with $N$ indicating the total number of different lighting conditions. No smoothness or albedo regularization is needed here.
\begin{equation}
	E(\bm{n},\rho) = \sum_k \big( \rho(p)  ({\bm{n}_p}^{T} A_k \bm{n}_p + b_k^{T} \bm{n}_p + c_k) - I_k(p) \big)^2
\end{equation}

The underlying principle of our enhancement method is based on the above photometric stereo theory, but we do not need to set the object to be static and manually change the lighting conditions; instead we have captured the RGB-D sequence of the object under arbitrary motion in uncalibrated natural illumination. In this case, the lighting variations induced by object motion resembles the classic photometric stereo in some way. We describe the derivations in the following.

Suppose we set the first frame as the reference frame, and also we can find the correspondences for pixels along the sequence. For example, for pixel $p$ in the reference frame, its correspondence in frame $k$ is $W(p)$. The appearance of the pixel $W(p)$ is generated as,
\begin{equation}
\begin{split}
	I_k(W(p)) &= \rho(p)  \big( (R_k \bm{n}_p)^{T} A (R_k \bm{n}_p) + b^{T} (R_k \bm{n}_p) + c \big) \\
	& = \rho(p)  \big( \bm{n}_p^{T} (R_k^{T} A R_k) \bm{n}_p + (b^{T} R_k) \bm{n}_p + c \big)
\end{split}
\end{equation}
where $\rho$ is the albedo for pixel $p$ which equals to the albedo of pixel $W(p)$ and $\bm{n}_p$ is surface normal under reference frame coordinate. $R_k$ is the rotation from the reference frame to frame $k$. Therefore, the surface normal for the corresponding pixel $W(p)$ in image $I_k$ can be represented as $R_k \bm{n}_p$. As demonstrated in the above equation, the rotation $R_k$ can be extracted and applied to the lighting vectors, from which we will get the lightings for frame $k$ as $R_k^{T} A R_k$, $b^{T} R_k$ and $c$. 

\begin{figure}[!h]
	\begin{center}
		\includegraphics[width=0.8\linewidth]{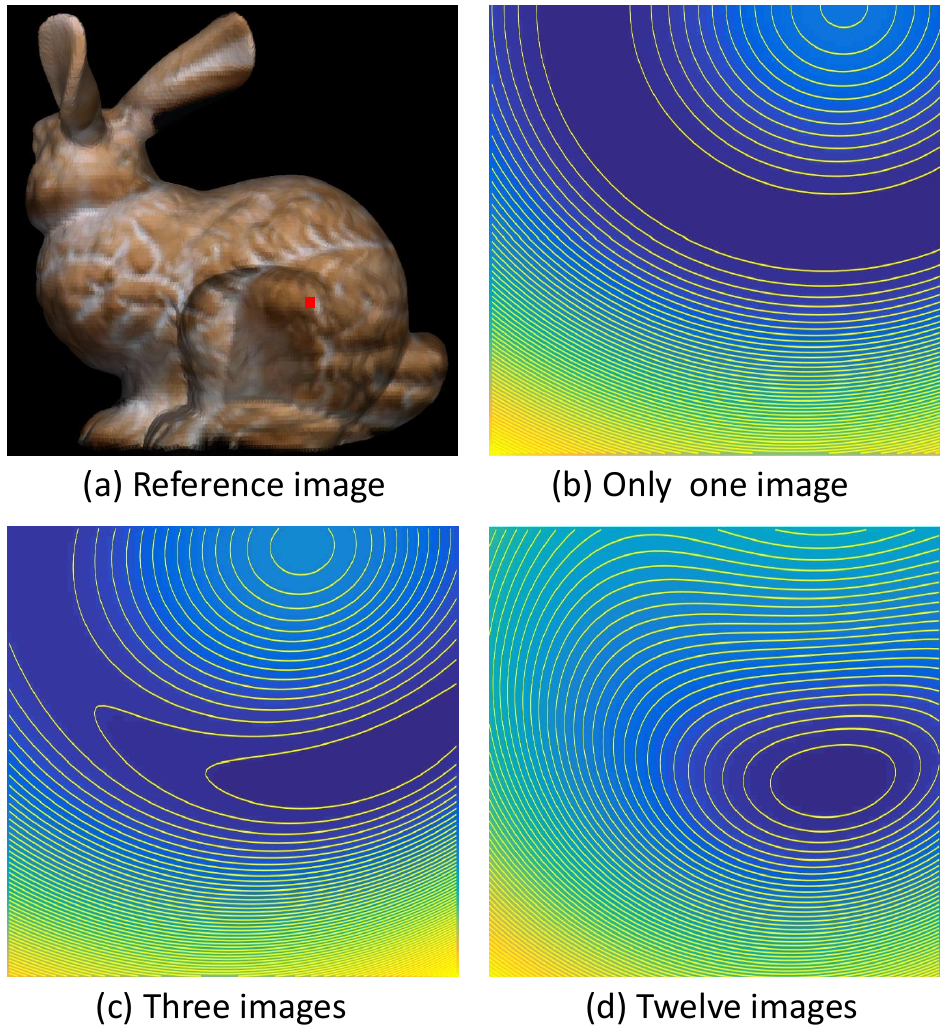}
	\end{center}
	\caption{Diverse rotation variations resolve local ambiguity. (a) shows a sampled image of the object with a reference pixel marked as red. In (b),(c) and (d) we plot the energy map for this reference pixel with x-axis and y-axis representing the two degree of freedom of a surface normal. The cooler color in these figures corresponds to smaller energy value. As we can see, given a single image the solution lies in a large band as shown in (b). With three images the normal converges better as shown in (c). And finally we will be able to find the optimal surface normal for the reference pixel if we have got enough images under rotation variations as shown in (d).}
	\label{Fig:energyPlot}
\end{figure}

Therefore as similar to the photometric stereo, the changes of lighting induced by the object motion provide valuable cues to recover the surface normal and its albedo. We illustrate this in Fig.~\ref{Fig:energyPlot} with the energy plot showing that with more and more images under diverse rotation variations included in the energy function, we are able to resolve the local ambiguity and converge to the optimal solution without relying on any smoothness regularizations.

%% file: pipeline.tex
\section{Pipeline}
\label{section3}

\begin{figure}[ht]
\begin{center}
   \includegraphics[width=0.9999\linewidth]{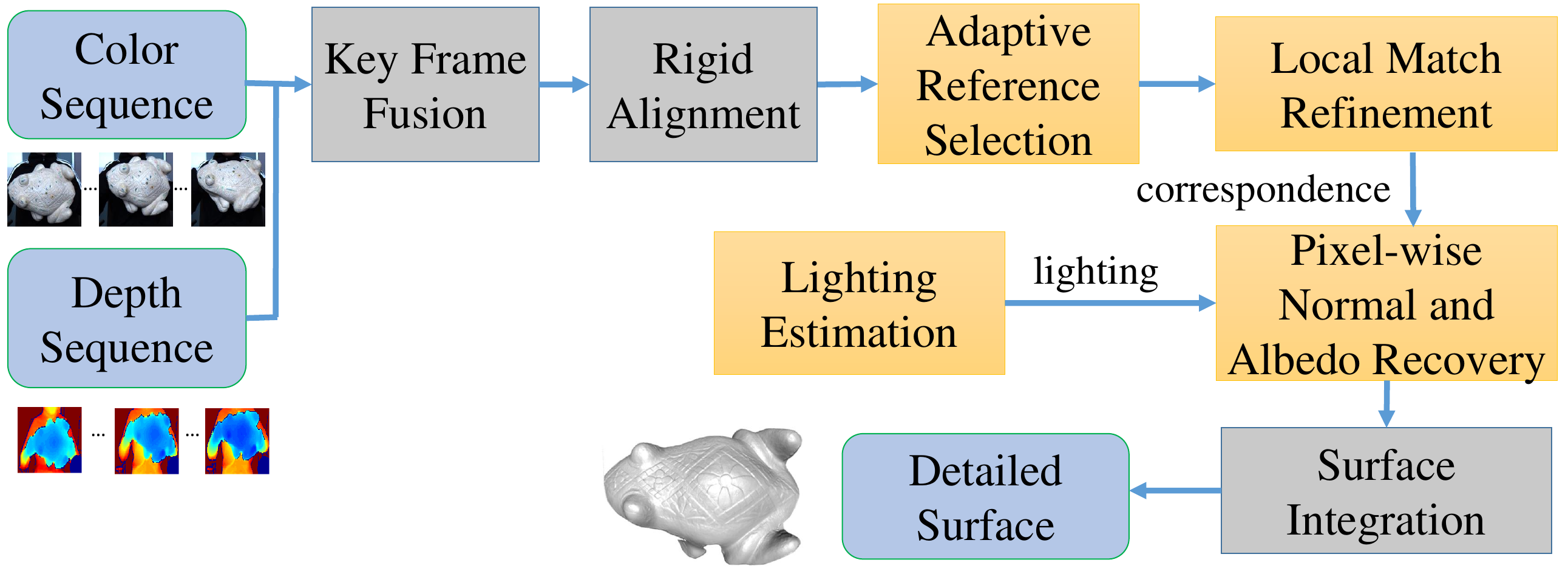}
\end{center}
   \caption{System Pipeline.}
   \label{Fig:pipeline}
\end{figure}

An overview of our depth enhancement and albedo recovery framework is shown in Fig.~\ref{Fig:pipeline}.

First, instead of using all those frames of the sequence which is redundant and computationally too expensive, we fuse every $M=20$ to generate $N$ key frame depth maps from the RGB-D sequence via KinectFusion~\cite{KinectFusion11}. They are smoother and more accurate than the raw depth maps. The extrinsic parameters between these key frames are computed and refined with bundle adjustment afterwards. Instead of using the first frame as the reference frame as in~\cite{Zuo17}, we propose an adaptive reference frame selection approach to find the best reference regions among all key frames. Next, a robust pixel matching strategy is proposed to find correspondences across the key frame images dealing with possible misalignments caused by imprecise initial depth maps or image distortion. We call this procedure Local Match Refinement, as the correspondences are locally searched and refined starting from the correspondences achieved by warping the images guided by the inital depth maps. For lighting estimation, we utilize the entire sequence to make the estimation more robust. Finally, given the computed lighting and correspondences along the sequence, we recover the surface normal and albedo image under our robust EM framework. To the end, the recovered normal can be integrated with any key frame depth map to generate a surface model with much more structural details revealed.

%% file: algorithm.tex
\section{Approach}
There are four major parts in the pipeline as shown in Fig.~\ref{Fig:pipeline}, including adaptive reference frame selection, searching correspondences among the images, lighting estimation, and normal and albedo recovery. We will describe them successively in details in the following sections. 

\subsection{Adaptive reference frame selection} 
\label{section40}
First of all, the key frame depth maps $D_1 \sim D_N$ are obtained via depth fusion with the corresponding color images denoted as $I_1 \sim I_N$. 


In the preliminary version~\cite{Zuo17}, the image of first key frame has been taken as the reference image and the correspondences are found from this reference image to other key frames. Theoretically speaking we can take any key frame as the reference one. However, it is not the case under real circumstances. First, as we rotate the object in front of the camera, we will get confront with the sampling issue. That is, for the same region on the object surface, the sampling rate is different across the images. When the surface region is closer to the camera and viewed more directly from the camera, more image details will get captured. This is different from the classic photometric stereo problem where the object keeps static and the sampling rate does not change across all the images. The overall resolution of the recovered surface normal will be subject to the reference image. Therefore, we want to select regions with higher sampling rate across all the key frames as the reference region. Second, there might be some undesirable artifacts on the first key frame such as cast shadow or specular reflections. Therefore, it is a futile attempt trying to find correspondences from the key frame with those artifacts to other frames which might result in incorrect correspondence. Taking all those factors into consideration, we believe it is necessary to select the best reference regions across all those key frames before the following matching procedure.

Technically, we formulate this as a labeling problem, that is we want to find out for every region on the object surface which key frame image is most appropriate to be selected as the reference image. 

To achieve this goal, first we obtain a fused 3D model with the KinectFusion~\cite{KinectFusion11} for the whole sequence. The correspondences between the fused model and every key frame model can be established during the fusion procedure. Next, we propose a labeling approach on the fused 3D model to select the reference regions adaptively. In details, suppose we have got $N$ key frames and the fused 3D model $S$. For each face $f$ on the 3D model $S$, we would like to assign a reference frame to it which is denoted as $l$ with $l\in(1,N)$. $\mathcal{L}$ represents the optimal label set with $\mathcal{L} = \{l_1,l_2...l_N\}$.

First, for each face $f$ we prefer the frame where it has the frontal view towards the camera so that more details can be captured in the image. We project each model face onto every key frame image $I_i$ and measure the area of projection region, which is related to the angle view proximity, image resolution and visibility constraint, as defined by,
\begin{equation}
E_d(l_f) = \exp \Big(-\frac{area[\Pi_l(f)]}{2\sigma_d^2} \Big),
\end{equation}

$E_d(l_f)$ indicates the cost of assigning label $l$ to the face $f$. $\Pi_l(f)$ is the projection of face $f$ to image $I_l$ and $area[\cdot]$ measures the image area after the projection.

To eliminate the effect of cast shadow or specularities in the key frame images, we penalize the frames where the intensity of corresponding image pixels is too dark or too bright. Instead of manually tuning the intensity threshold, we fit a Gaussian model to the corresponding pixels across all the key frames. That is, for each model face $f$ we project onto each key frame and compute the average intensity of pixels inside its projection. We collect all those intensities and get a Gaussian model with its mean denoted as $\mu_f$ and variance as $\sigma_f$. Then we define the cost as,

\begin{equation}
E_h(l_f) = \exp \Big(-\frac{(I_l(\Pi_l(f))-\mu_f)^2}{2\sigma_f^2} \Big)
\end{equation}

Besides, we add a smoothness term defined after the Potts Model~\cite{Pottsmodel} to encourage the same label for neighboring faces and prevent too many small label fragments. 
\begin{equation}
E_s(\mathcal{L}) = \sum_{(f,g)\in \mathcal{N}} [l_f \neq l_g],
\end{equation}
where $\mathcal{N}$ is the neighboring set and face $f$ and $g$ shares the same edge. 

Finally, we minimize the following energy function~\cite{kolmogorov2015new} to get the label assignments for every face of the 3D model.
\begin{equation}
E(\mathcal{L}) = \lambda_d \sum_{f \in S} E_d(l_f) + \lambda_h \sum_{f \in S} E_h(l_f) + E_s(\mathcal{L})
\end{equation}

\begin{figure}[!ht]
	\begin{center}
		\includegraphics[width=0.9999\linewidth]{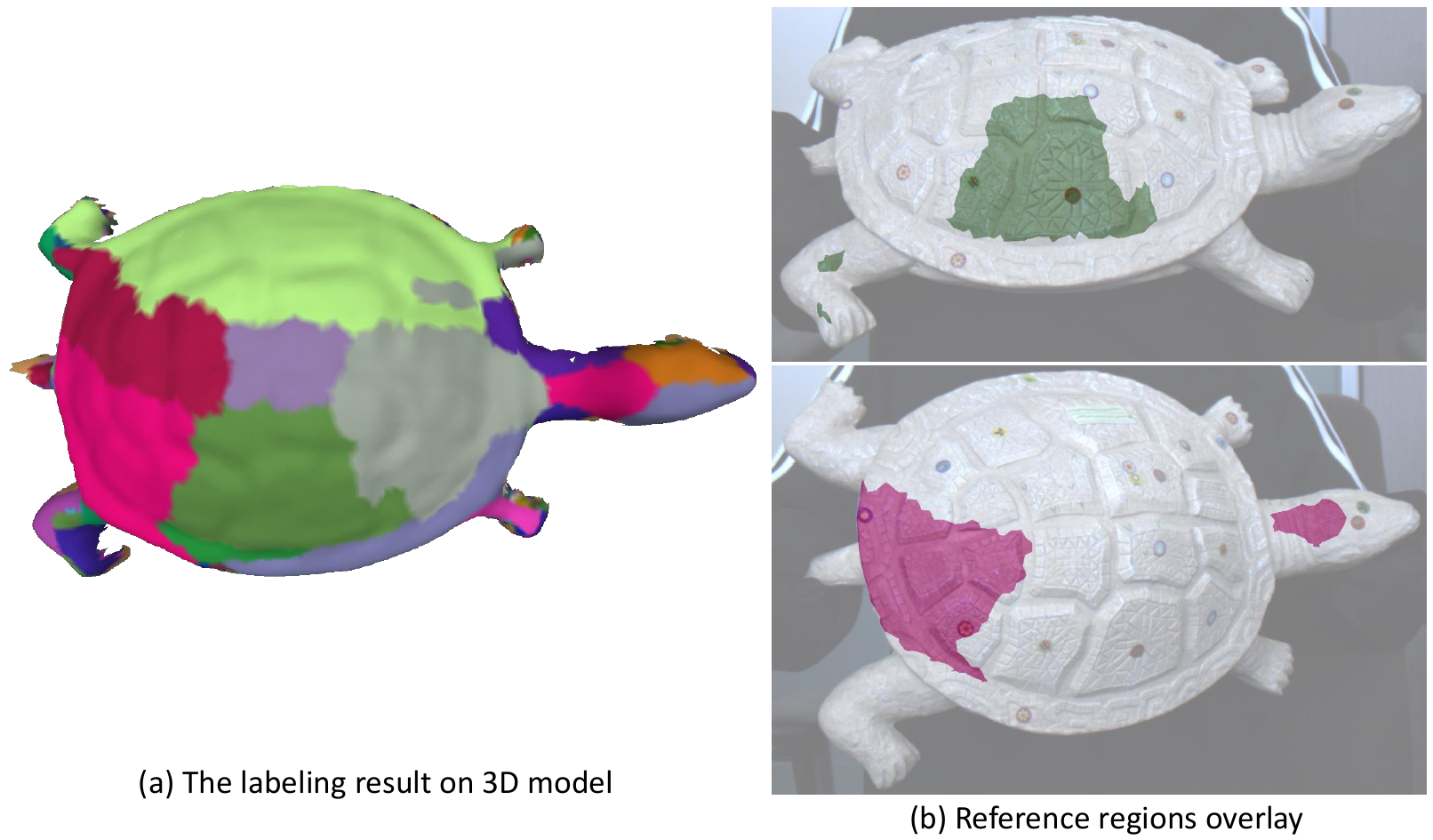}
	\end{center}
	\caption{Results on adaptive reference frame selection. (a) shows the labeling results on the 3D model with different colors representing different key frames. (b) shows the labeling results on two sampled key frame images by projecting the labeled 3D model onto those corresponding frames.}
	\label{Fig:selection}
\end{figure}

We illustrate the labeling result in Fig.~\ref{Fig:selection}.

After getting the labeled results on 3D model, we project it onto every key frame image since the following pixel matching and normal computation procedures are both performed in the image domain. Afterwards we will get the reference regions selected adaptively in those key frame images, as shown in Fig.~\ref{Fig:selection}(b).


\subsection{Robust Pixel Matching} 
\label{section41}


\subsubsection{Rigid alignment} 
\label{section411}

First, the global rigid transformation between key frames are calculated by detecting SIFT or ORB features followed by feature matching. These extrinsic parameters are further refined with bundle adjustment and finally we get the rotation $R_1 \sim R_N$ and translation matrix $T_1 \sim T_N$ with respect to a global coordinate for each key frame from which we can warp any key frame to other frames.

\subsubsection{Lighting insensitive Local Match Refinement} 
\label{section412}

\begin{figure*}[!]
	\begin{center}
		\includegraphics[width=0.9999\linewidth]{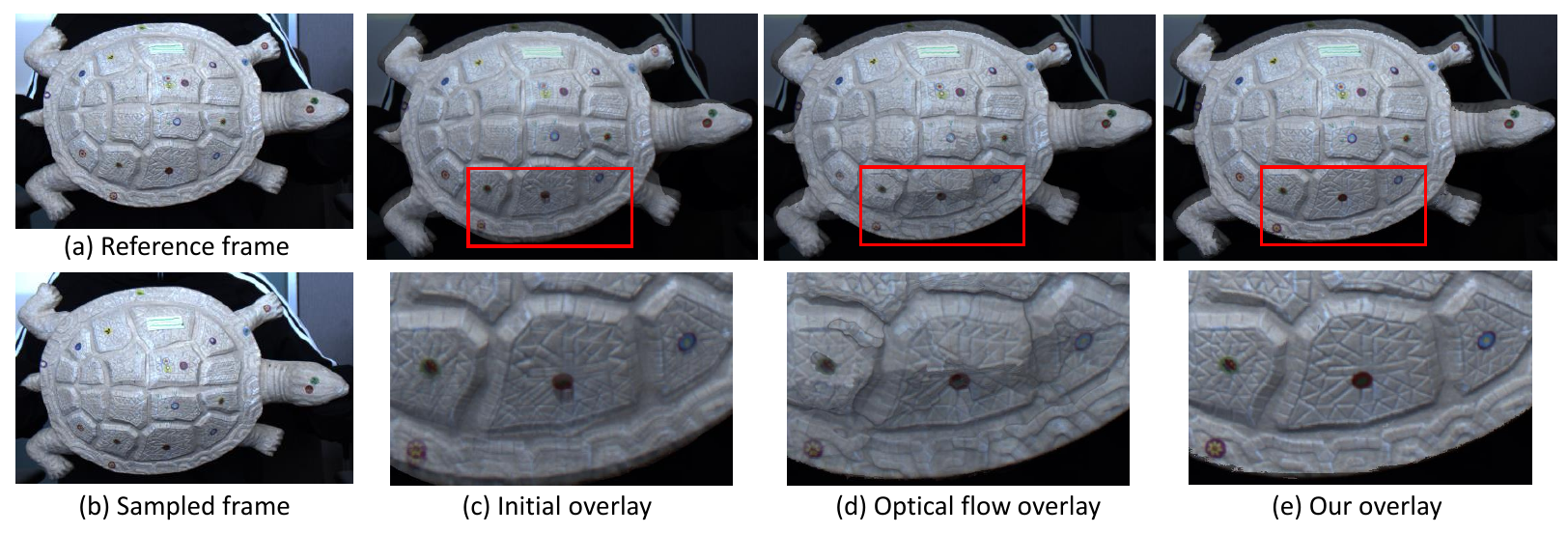}
	\end{center}
	\caption{Demonstration of correspondence matching. (a) is the reference frame and (b) is one sampled key frame. Image(b) is warped to the reference frame with current transformation, and (c) displays the warped image overlaid with image(a). (d) shows the overlaid result using the flow map computed from warped image and the reference image~\cite{Brox11}. (e) is the overlaid image after applying our proposed lighting insensitive robust matching.}
	\label{Fig:flow}
\end{figure*}

These key frames can be warped into any reference frame given the current transformation. However misalignments still exist after bundle adjustment as shown in Fig.~\ref{Fig:flow}(c), which is caused by the imprecise depth maps, image distortion and the imperfect synchronization between the captured color and depth sequence. Optical flow is often used as a solution to find correspondences between two images. Considering that the misalignment may be severe, we have tried to use a large displacement optical flow computation approach~\cite{Brox11} to find the correspondences between the warped image and the reference image. However, since the consistency assumption is not maintained in our case, the alignment has got even worse in some part with great illumination changes as displayed in Fig.~\ref{Fig:flow}(d). Besides, since the reference regions have been selected adaptively across the sequence, we want to find the correspondences from those reference regions to other frames rather than performing the optical flow computation in the whole image from any frame to other frames. Therefore, we develop a robust matching method to deal with these issues.


Our matching approach is implemented on every reference region and we find correspondences from the reference region to other key frame images. Suppose we have the reference depth map (region) and corresponding color image denoted as $D_{\text{ref}}$ and $I_{\text{ref}}$ respectively. In this paper, depth means the distance to the image plane. For each pixel $p=(u,v)$ in $I_{\text{ref}}$, its current corresponding pixel $q$ after bundle adjustment in image $I_k$ is computed as,
\begin{equation}	
\lambda \left [ \begin{array}{c}
   		 q  \\
   		 1 \\
		\end{array}\right]  =K \Bigg( R_{k} \big(R_{\text{ref}}^{-1} (K^{-1}  \left [ \begin{array}{c}
   		 u  \\
   		 v \\
  		 D_{\text{ref}}(u,v) \\
		\end{array}\right] - T_{\text{ref}})\big) + T_k \Bigg)
\end{equation}

In the above equation, $K$ is the camera intrinsic matrix. $R_{ref}$ and $T_{ref}$ are the rotation and translation matrix that transform the 3D point from world coordinate to the reference frame. Similarly, $R_k$ and $T_k$ are the rotation and translation matrix for frame $k$.

The corresponding pixel $p$ in $I_{\text{ref}}$ and $q$ in $I_k$ may not be the correct correspondence because of the misalignment. Therefore, we implement a local search strategy to find its best matching pixel in $I_k$.

For each pixel $p$ in $I_{\text{ref}}$, we set a searching region around it and find its best match in $I_k$ via NCC (Normalized Cross Correlation). However, the intensity consistency is not preserved as the object is subject to arbitrary movements. This makes the original NCC not suitable for matching in this case. To deal with this problem, we apply chromacity normalization in the color image to eliminate the effect of lighting variations~\cite{Finlayson03} and use the normalized images for matching.
For each pixel $p$, its appearance is generated as,
\begin{equation}
	I_{ch}(p) = \rho_{ch}(p) s(\bm{n}_p)  \quad   ch\in\{R,G,B\},
\end{equation}
in which $s(p)$ is the shading function that accounts for the lighting or normal variation.

So the chromacity normalization is implemented as,
\begin{equation}
	I_{ch}^{cn}(p) = \frac{I_{ch}(p)}{I_{R}(p) + I_{G}(p) + I_{B}(p)}  \quad  ch\in\{R,G,B\}
\end{equation}

After the above normalization, NCC can then be applied for the matching which will be insensitive to the photometric inconsistency induced by lighting factor. For the NCC computation, we perform it separately on each channel of the color image.

Specially, the color image $I_{\text{ref}}$ is warped to the color frame $I_k$ under the guidance of $D_{\text{ref}}$ and we get the warped color image $I_{\text{ref}_k}$. The NCC patch matching is implemented in $I_{\text{ref}_k}$ with $I_k$ instead of using $I_{\text{ref}}$ directly. Since $I_k$ and $I_{\text{ref}_k}$ are in the same viewpoint, the fattening effect of NCC is successfully avoided.

Although for each pixel in $I_{\text{ref}}$ (or $I_{\text{ref}_k}$) we can find the corresponding pixel in $I_k$ that has the largest matching score, we cannot guarantee they are always the correct correspondence. To tackle this problem, we only keep the pixels that are reliable and use these pixels as control vertices to deform all the other pixels to find their correct correspondences.

Our criteria of reliable matches is that, 1) the largest matching score should be larger than $\text{thres}_S$; 2) the difference between the largest score and second largest score of local peaks should be larger than $\text{thres}_{\Delta}$. If these principles are maintained, the pixel in the searching region that has the largest score is chosen as the correspondence. $\text{thres}_S$ is set to be 0.75 and $\text{thres}_{\Delta}$ is 0.05 in this paper for all the experiments.

Next we use these reliable matches as control vertices to deform the image $I_{\text{ref}_k}$ so that it has an optimal match with $I_k$.
For each control vertices $o_{l}$ in $I_{\text{ref}_k}$, the deformation function is defined as
\begin{equation}
	f(o_{l}) = o_{l} + \Delta_{l},
\end{equation}
where $\Delta_{l}$ is the motion vector between the optimal correspondence and its initial correspondence in $I_k$.

For other pixels the deformation is formulated via bilinear interpolation with control vertices~\cite{Zhou2014color},
\begin{equation}
	f(u) = u + \sum_{l} (\theta_{l}^{u} \Delta_{l}),
\end{equation}

The interpolation coefficients $\theta_{l}^{u}$ is set according to the distance to control vertices and only neighboring vertices will affect the deformation.

Finally, our objective function is defined to maintain the photo consistency of the two normalized images.
\begin{equation}
	E(\Delta) = \sum_{p}\Big( I_{\text{ref}_k}^{cn}(f\big(p,\Delta)\big) - I_{k}^{cn}(p) \Big) + \lambda \sum_{l}||\Delta_{l} - \hat{\Delta}_{l} ||^2 ,
\end{equation}

where $\hat{\Delta}$ is the initial deformation vector for the control vertices between the current optimal correspondence obtained from matching and its initial correspondence. $\lambda$ is the control weight set to be 10 in this paper. Since we have good initials $\hat{\Delta}$, the optimization will converge quite fast.

Some matching results are demonstrated in Fig.~\ref{Fig:flow}(e). While our matching approach is implemented on each reference region, it does not stop us from performing it on the whole image. For better demonstration, we show the matching results in Fig.~\ref{Fig:flow} with one particular frame taken as the reference image.

\subsection{Lighting estimation} 
\label{section42}

In this section, we demonstrate how to compute the lighting parameters $A$, $b$, $c$ for each reference region. Since The unknown albedo poses challenges for lighting estimation, there are some methods that cluster the image into different parts and use the mean value as their albedos. The lighting and albedo is estimated in an iterative way. Instead of trying to resolve the ambiguity from a single frame, in this paper, we employ the aligned color sequence and depth maps for robust lighting estimation, eliminating the need to make prior assumptions about albedo.

With the aligned color images we can compute the ratio images with respect to the reference image region, from which the albedo will get canceled out. In details, for each pixel $p$ in $I_{\text{ref}}$, suppose its corresponding pixel in $I_k$ is denoted as $q$, then the ratio value is computed as,
\begin{equation}
\begin{split}
	\frac{I_{k}(q)}{I_{ref}(p) } & = \frac{\rho(q)  ( {\bm{n}_q}^{T} A {\bm{n}_q} + b^{T} {\bm{n}_q} + c ) }{\rho(p) ( {\bm{n}_p}^{T} A {\bm{n}_p} + b^{T} {\bm{n}_p} + c ) } \\
	 &= \frac{  {\bm{n}_q}^{T} A {\bm{n}_q} + b^{T} {\bm{n}_q} + c }{ {\bm{n}_p}^{T} A {\bm{n}_p} + b^{T} {\bm{n}_p} + c }
\end{split}
\end{equation}

Therefore, the environmental lighting can be achieved from the following minimization,
\begin{equation}
	\mathop{\arg\min}_{A,b,c} \sum_{k} \sum_{p\in I_{\text{ref}}} \gamma_p ( \frac{  {\bm{n}_q}^{T} A {\bm{n}_q} + b^{T} {\bm{n}_q} + c }{ {\bm{n}_p}^{T} A {\bm{n}_p} + b^{T} {\bm{n}_p} + c } - \frac{I_{k}(q)}{I_{\text{ref}}(p) } )^2 ,
\end{equation}

The normal $\bm{n}$ are approximated using the initial normals computed with the key frame depth maps. The weighting term $\gamma_p$ is set to prevent using pixels with intensity that are too dark or too bright which might be caused by cast shadow or specularities. Besides, we also ignore pixels with great image gradients that are sensitive to misalignments. The lighting vectors can get updated iteratively after the albedo recovery with refined normal maps.

\subsection{Normal and albedo recovery} 
\label{section43}

With the key frame color images all aligned into the reference image regions ($I_1^W \sim I_N^W$), and the estimated environmental lighting $(A,b,c)$, we are ready to recover the surface normal and its albedo. We have the object rotation matrix $R_1 \sim R_N$ for each frame with respect to the reference frame. Then for each pixel $p$ in the reference frame, our goal is to find the optimal albedo $\rho(p)$ and normal $\bm{n}(p)$ conforming the pixel observations $\bm{I}(p)=\{I_k^W(p)\}_{k=1}^N$. \textit{We drop the index of pixel locations for simplicity in the following description.} The objective function can be defined as:
\begin{equation}
	E(\bm{n},\rho| \bm{I}) = \sum_{k} \big(  s_k(\bm{n})\rho - I_k^W \big)^2,
\end{equation}
\begin{equation}
	s_k(\bm{n}) = \bm{n}^{T} R_k^{T} A R_k \bm{n} + b^{T} R_k \bm{n} + c
\end{equation}

The surface normal and albedo can be estimated from minimization of the above function. However, the outliers have not been taken into consideration. They will affect the result if the observations violate the Lambertian assumption. To deal with these outliers, we introduce a set of hidden states $H_k=\{0,1\}$ indicating whether the observation is actually generated by the Lambertian model. The expectation-maximization (EM) algorithm is exploited to solve the problem. While our formulation is inspired by~\cite{Wu10}, we extend it from its original directional light assumption to general lighting. More specifically, we denote the parameters to be estimated as $\Omega=\{\bm{n},\rho,\sigma,\alpha\}$ and the observation probability conditioned on parameters $\Omega$ is given as,
\begin{equation}
\begin{split}
	P(I_k^W | \Omega )& = \alpha  \frac{1}{\sqrt{2\pi} \sigma} \exp \Big(- \frac {(   s_k(\bm{n}) \rho  - I_k^W ) ^2} {2\sigma^2} \Big)  \\
 				&+ (1-\alpha) \frac{1}{C}
\end{split}
\end{equation}

$P(H_k=1) = \alpha$ is the prior probability of $H_k$ indicating the proportion of observations generated by the Lambertian model. $\frac{1}{C}$ is the probability as being an outlier which is assumed to be uniform distribution and we set $C$ to be 10 in our implementation.

The posterior probability of the hidden variable $H_k$ is updated in every E-step using the following equation given the computed parameters $\Omega'$ in current iteration and the observation $I_k^W$,
\begin{equation}
\begin{split}
	\omega_k &= P(H_k=1 | I_k^W,\Omega') \\
			&= \frac{\alpha  \exp (- \frac {(   s_k(\bm{n})\rho  - I_k^W )^2} {2\sigma^2}) }{\alpha  \exp (- \frac {(   s_k(\bm{n})\rho  - I_k^W )^2} {2\sigma^2}) +  \frac{1 - \alpha}{C} }
\end{split}
\end{equation}

Next, in the following M-step, we maximize the complete-data log-likelihood given the marginal distribution $H_k$ obtained from the E-step.
\begin{equation}
\begin{split}
P(\Omega) &= \sum_k \log P(I_k^W, H_k=1  | \Omega) \omega_k \\
	     &+ \sum_k \log P(I_k^W, H_k=0 | \Omega)(1- \omega_k) \\
		&= \sum_k \log \Big(\frac{\alpha}{\sqrt{2\pi}\sigma} \exp \big(- \frac {(s_k(\bm{n} \big) \rho  - I_k^W )^2} {2\sigma^2}\big) \Big) \omega_k \\
		&+\sum_k  \log \Big(\frac{1-\alpha}{C} \Big)(1-\omega_k)
\end{split}
\end{equation}

To maximize the above function, we set the first derivative of $P$ with respect to $\alpha$, $\sigma$ and $\rho$ equal to zero. In this way, the updating rules for these parameters are obtained,
\begin{equation}
\begin{split}
	\alpha &= \frac{1}{N} \sum_k \omega_k \\
	\sigma  & = \sqrt{ \frac{\sum_k (s_k(\bm{n})\rho  - I_k^W )^2 \omega_k}{\sum_k \omega_k} } \\
	\rho & = \frac{1}{\sum_k s_k(\bm{n})^2 \omega_k} \sum_k s_k(\bm{n})  \omega_k  I_k^W \\
\end{split}
\end{equation}

Since the function $P$ is nonlinear to surface normal $\bm{n}$, the updated normal is achieved by fixing other parameters and solving the following energy minimization.
\begin{equation}
	\mathop{\arg\min}_{\bm{n}}  \sum_k \big( (\bm{n}^{T} R_k^{T} A R_k \bm{n} + b^{T} R_k \bm{n} + c)\rho - I_k^W \big)^2 \omega_k
\end{equation}

The above EM iterative optimization process is performed until no further improvement on the recovered normal and albedo. The initial parameter of $\alpha$ and $\sigma$ is set to be 0.75 and 0.05 respectively for all the datasets used in this paper.

Finally, the recovered normal is all warped to some key frame and get integrated with the depth map to get enhanced surface geometry with structural details~\cite{Zhang12}.

\subsection{Implementation details} 
\label{section44}

As a preprocessing step, the object is first segmented from the image by integrating both color and depth information into GrabCut~\cite{Rother04} framework. We manually masked the first frame with the rest of frames segmented automatically.  

We implement most parts of our framework in Matlab on a desktop with 8-core 3.6 GHz Intel CPU and 32 GB memory and it takes us approximately 820s to process a dataset with $500 \sim 600$ frames. Considering that the normal and albedo is computed in pixel-wise manner, the running time could be reduced further with parallel computation.

%% file: exp.tex
\section{Experimental results}
In the experiments, we validate our method on synthetic and real datasets with quantitative and qualitative evaluation.

\subsection{Synthetic datasets} 
\label{section51}

In this section we perform quantitative evaluations of our method on several synthetic models. First given the 3D model, twenty images together with their corresponding depth maps are rendered under natural illumination. The rendered ground-truth depth maps are over smoothed to filter out the structural details. Those smoothed depth maps and rendered color images are taken as input for our method. We have compared our method with a shading refinement approach ~\cite{Or-el15} from a single RGB-D image which has achieved good performance on depth refinement.

\begin{figure}[!h]
	\begin{center}
		\includegraphics[width=0.9999\linewidth]{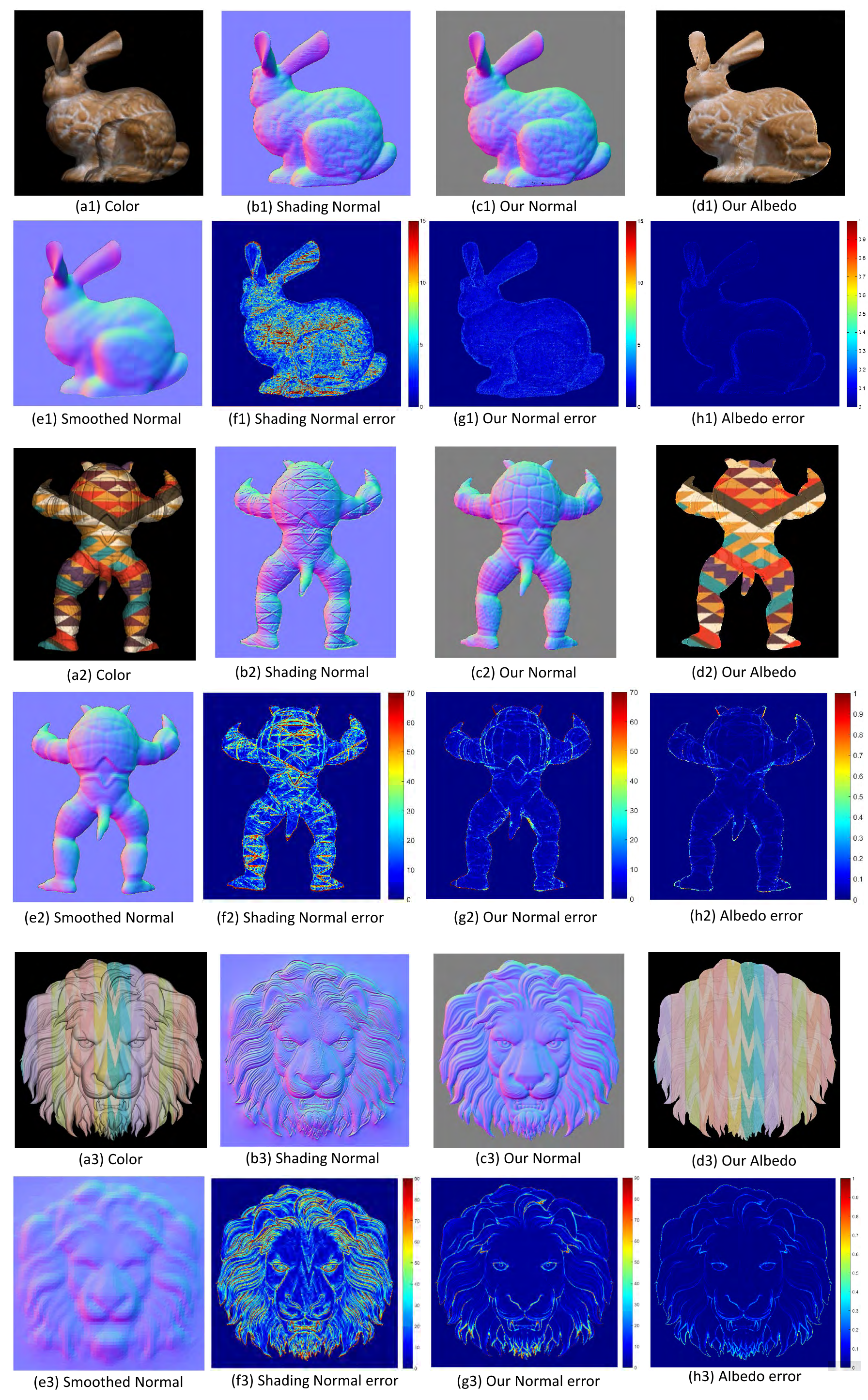}
	\end{center}
	\caption{Results on synthetic models. (a1-a3) is the rendered color image of the reference frame; (e1-e3) shows the normal map of the ground-truth mesh after over smoothing; (b1-b3) is the normal map computed after applying shading refinement on the reference frame with its error map displayed in (f1-f3); (c1-c3) and (g1-g3) are the normal map and its corresponding error map achieved by our method. Our recovered albedo map and its error map is also demonstrated in (d1-d3) and (h1-h3) respectively.}
	\label{Fig:bunny}
\end{figure}

Fig.~\ref{Fig:bunny} shows the comparison results of our recovered normal map and surface. For each model, the first column is the reference color map and over smoothed mesh (displayed as normal map). These are the input for the shading refinement method~\cite{Or-el15}. The output of the shading refinement method is displayed in the second column. The texture copy artifacts are caused by imperfect separation of albedo and shading layers. In comparison, the surface normal can be recovered successfully with our pixel-wise recovery method with quite small error shown in the third column. The albedo map computed from our method together with its error map is demonstrated in the last column.

We display the quantitative results in Table~\ref{Table:error} showing the mean error of computed normal maps, extracted albedo images and also the enhanced depth maps. As we can see, the error of our computed normal map is quite small as compared with the shading refinement approach. For the Armadillo and Lion model, the normal error of the shading refinement approach becomes even larger than the initial over smoothed normal as caused by the texture copy problem. We have also performed evaluation on our recovered albedo image with the mean error shown in the Table~\ref{Table:error}. We have normalized the images into 0~1. Since we cannot resolve the scale ambiguity of the computed albedo and shading image, we have calculated a scale factor with six randomly selected pixels in the ground-truth albedo image, which are divided by the values of corresponding pixels in our recovered albedo image.

\begin{table}[h]
\centering
	\caption {Quantitative Evaluation.} 
	\label{Table:error} 
	\begin{center}
		\begin{tabular}{ | c | c | c | c | }
			\hline
			Model & \textbf{Bunny} & \textbf{Armadillo}  & \textbf{Lion} \\ \hline
			Initial depth error  &  1.2150 & 1.1129 & 1.3927\\ \hline
			Initial normal error (degree)  & 8.4236 & 11.7554 & 14.3837\\ \hline
			Shading normal error (degree)  & 6.7591 & 14.6649 & 23.5301\\ \hline
			Our normal error (degree)  & \textbf{1.3485} & \textbf{4.5130} & \textbf{6.1094}\\ \hline
			Our albedo error  & 0.0127 & 0.0384 & 0.0295\\ \hline
			Our depth error  &  0.2506 & 0.3129 & 0.2927\\ \hline
		\end{tabular}
	\end{center}
\end{table}

\begin{figure}[!h]
	\begin{center}
		\includegraphics[width=0.9999\linewidth]{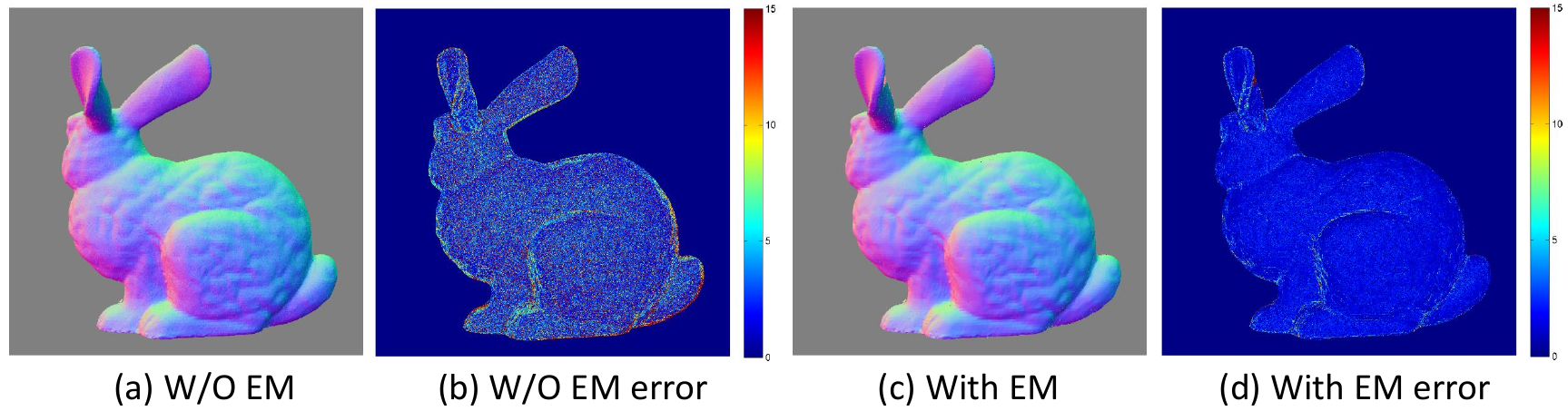}
	\end{center}
	\caption{Results when adding salt and pepper noise. (a) shows the computed normal map without our EM framework and (b) is its error map; The normal and error map after applying our EM optimization are shown in(c) and (d) respectively.}
	\label{Fig:bunnyEM}
\end{figure}

Fig.~\ref{Fig:bunnyEM} is shown to demonstrate the effectiveness of our EM framework for robust normal recovery in the presence of outliers. We have picked four out of those twenty images randomly and added the salt and pepper noise with 0.50 density. It means the abrupt noise will affect approximately fifty percent of the image pixels. As we can see from the first two columns, the recovered normal map without EM optimization is noisy (the mean error is 8.37 degrees), while we can achieve much better performance after applying our EM method, which is shown in the last two columns and the mean error decreased to 1.49 degrees.

\begin{figure}[!h]
	\begin{center}
		\includegraphics[width=0.9999\linewidth]{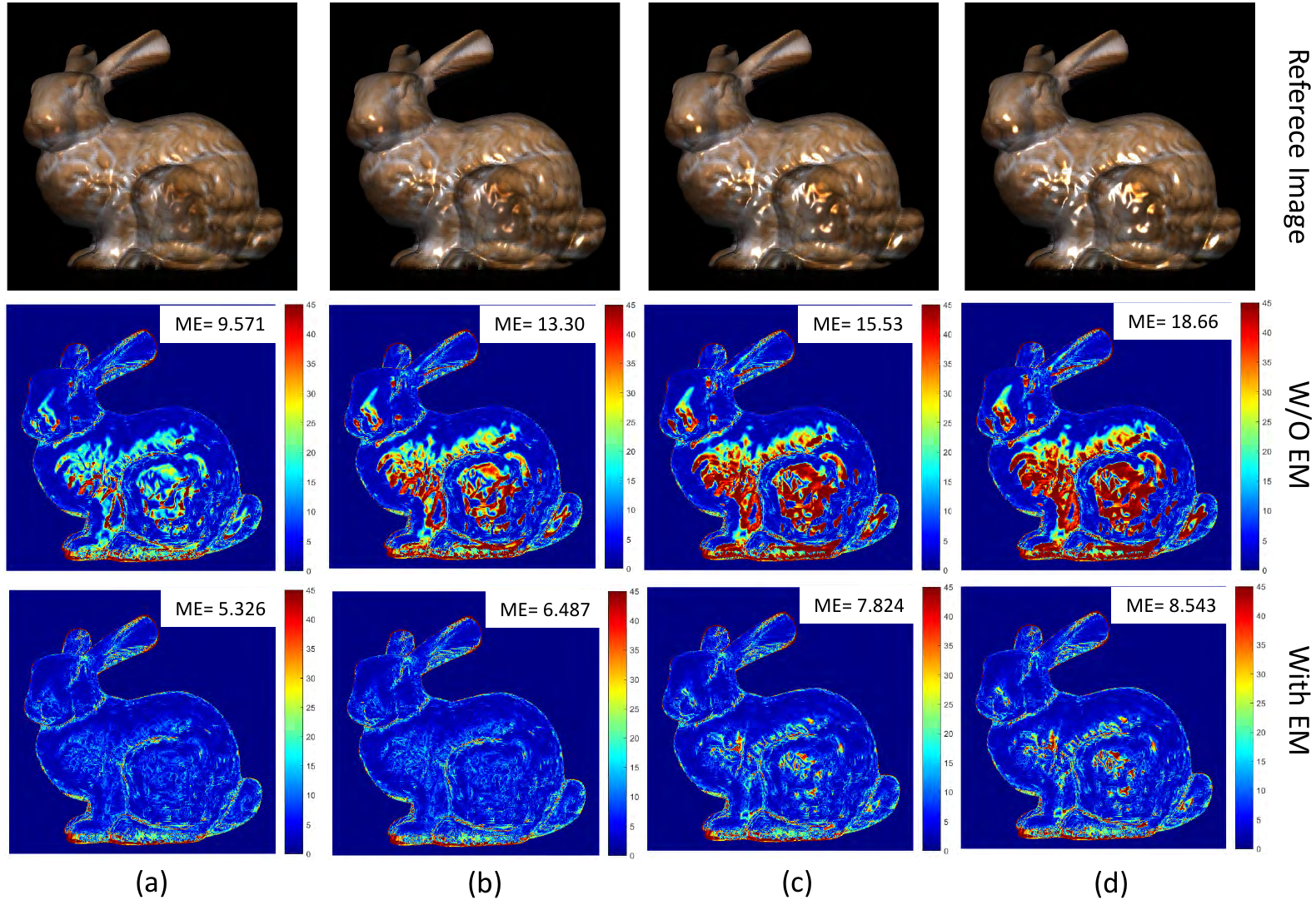}
	\end{center}
	\caption{Results on objects with specular reflection. We show the reference images on the first row. The specular proportion increases from left to right. The second row are the corresponding error map of the recovered surface normal without applying our EM optimization. The last row shows the error map of the recovered surface normal from our approach. The mean error (in degree) of the estimated surface normal is displayed at the right upper corner of each error map.}
	\label{Fig:bunnySpecular}
\end{figure}

To further test our method on handling objects with imperfect Lambertian reflection such as specularity, we have used the Standford Bunny model and synthesized images with specular component in addition to the diffuse part. We divide the specular component by its corresponding pixel value of the synthesized image, and we get the percentage of specular component with respect to the whole image. We use the average number as a measurement for the proportion of specularity. As shown in Fig. ~\ref{Fig:bunnySpecular}, the error of the computed surface normal increases as we enlarge the proportion of specularity. But we can still get reasonable results since we have handled the outliers explicitly in our EM based normal recovery approach. As we can see in the second row of Fig.~\ref{Fig:bunnySpecular}, without applying our EM based optimization framework, the error is quite large for the recovered normal map especially in regions showing high specularity, which demonstrates that the proposed EM based approach is essential to handle outliers.

\begin{figure}[h!]
	\begin{center}
		\includegraphics[width=0.9999\linewidth]{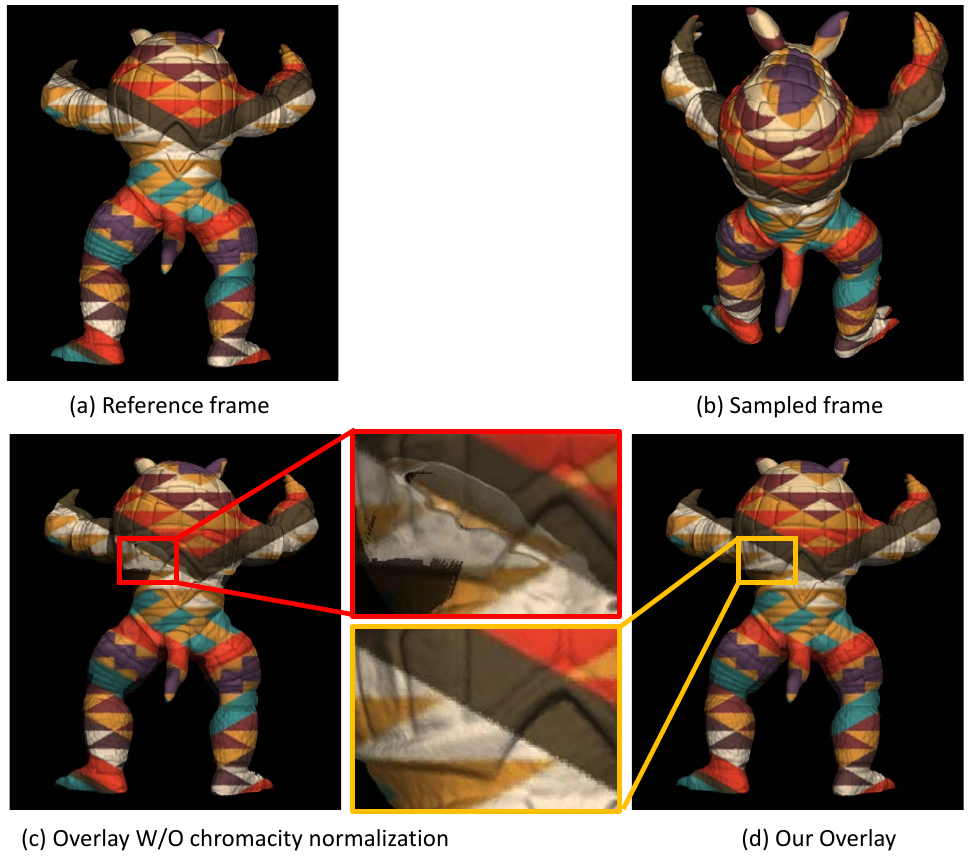}
	\end{center}
	\caption{Evaluation of our lighting insensitive local match refinement method. The reference frame and one sampled frame are shown in (a) and (b) respectively. We warp the sampled image back to the reference frame with the computed correspondences. (c) shows the overlay of the reference image and the warped image without applying the chromaticity normalization. (d) is the overlay of the warped image achieved from our robust matching approach.}
	\label{Fig:CN_validation}
\end{figure}

We have also evaluated the performance of our lighting insensitive local match refinement method on the synthetic datasets as we have the ground-truth pixel correspondences. From each synthetic dataset, we have selected a frame that has relatively large rotation angle with respect to the reference frame. The mean error of the estimated matches are shown in Table~\ref{Table:NC:error} with. The error is represented in pixel. As shown in Table~\ref{Table:NC:error}, without our chromaticity normalization procedure, the pixel error becomes even larger than the original ones computed by image warping using the initial depth maps. On the contrary, we can get the correspondences with the mean error less than one pixel with our lighting insensitive local match refinement. We also demonstrate the alignment in Fig.~\ref{Fig:CN_validation} with the reference image overlaid with the selected frame warped back to the reference frame.

\begin{table}[h]
	\centering
	\caption {Quantitative Evaluation of the alignment (unit: pixel error).} 
	\label{Table:NC:error} 
	\begin{tabular}{ | c | c | c | c | }
		\hline
		Model & \textbf{Bunny} & \textbf{Armadillo}  & \textbf{Lion} \\ \hline
		Initial pixel error  & 3.24 & 3.86 & 5.74\\ \hline
		\tabincell{c}{Matching without \\chromacity normalization} & 5.85 & 7.46 & 6.42\\ \hline
		Ours  & 0.65 & 0.72 & 0.86\\ \hline
	\end{tabular}
\end{table}

\subsection{Real datasets} 
\label{section52}

We have captured the datasets of real objects using the depth sensor of Kinect V2 with resolution of $512 \times 424$ and a PointGrey color camera with resolution of $1920 \times 1080$. Several objects are captured, namely the Frog, Shoe, Backpack, Turtle, etc. We will demonstrate the comparison results of our recovered surface normal and albedo with some state-of-the-art approaches in the following.

\subsubsection{Surface normal and geometry recovery}

\begin{figure*}[!]
	\begin{center}
		\includegraphics[width=0.95\linewidth]{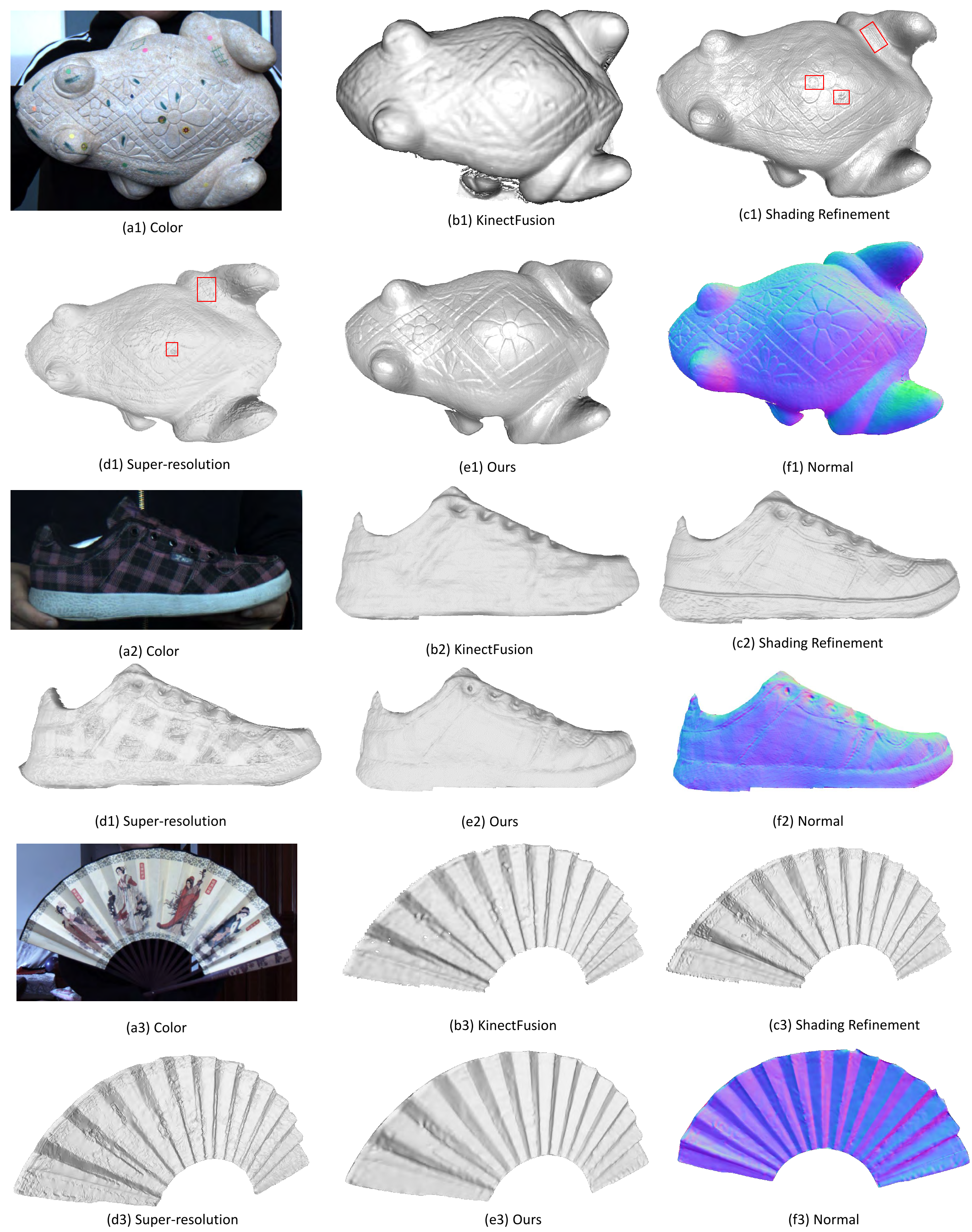}
	\end{center}
	\caption{Comparison results on Frog, Shoe and Chinese Fan model. (a1) (a2) and (a3) are the reference color images of Frog, Shoe and Chinese Fan respectively. The outputs from KinectFusion are shown in (b1) (b2) and (b3). The results computed by shading refinement method~\cite{Or-el15} are displayed in (c1) (c2) and (c3). (d1) (d2) and (d3) are the meshes computed by depth super-resolution method~\cite{Haefner18}. Finally, (e1) (e2) and (e3) are the output meshes from our approach with their corresponding normal maps displayed in (f1) (f2) and (f3).}
	\label{Fig:frogshoefan}
\end{figure*}

Fig.~\ref{Fig:frogshoefan} shows the comparison results of a Frog, Shoe and Chinese Fan model. We have made comparisons with a shading refinement approach~\cite{Or-el15} and a depth super-resolution approach~\cite{Haefner18} which deals with depth super-resolution and shading refinement problems simultaneously in an unified framework. The color images shown in Fig.~\ref{Fig:frogshoefan}(a1)(a2)(a3) and their corresponding depth images are taken as input for those two methods. As we can see from Fig.~\ref{Fig:frogshoefan}(b1)(b2)(b3) the surface has got over-smoothed after the fusion~\cite{KinectFusion11} and the small surface details cannot get revealed as restricted by the resolution and accuracy of the Kinect depth sensor. The shading refinement approach~\cite{Or-el15} is able to recover some surface details, but some textures are hallucinated as geometry details as well (Fig.~\ref{Fig:frogshoefan}(c1) (c2) and (c3)).
Fig.~\ref{Fig:frogshoefan}(d1) (d2) and (d3) displays the results of depth super-resolution~\cite{Haefner18} for which the colorful textures have caused unpleasant artifacts on the recovered surface as they have also assumed that the surface albedo is piecewise constant.
Fig.~\ref{Fig:frogshoefan}(e1,e2,e3) and Fig.~\ref{Fig:frogshoefan}(f1,f2,f3) displays our final results of recovered meshes and surface normals. For the Frog and Shoe model, the small surface details have got successfully extracted and revealed in our results without affected by the textures. For the Chinese Fan model, it contains some concave parts which could cause cast shadow on the images. Although it does not have much small geometric details, the pleats become more sharp in our recovered mesh as compared to the fusion results with smooth surface on other parts as it should be.


\begin{figure*}[!]
	\begin{center}
		\includegraphics[width=0.965\linewidth]{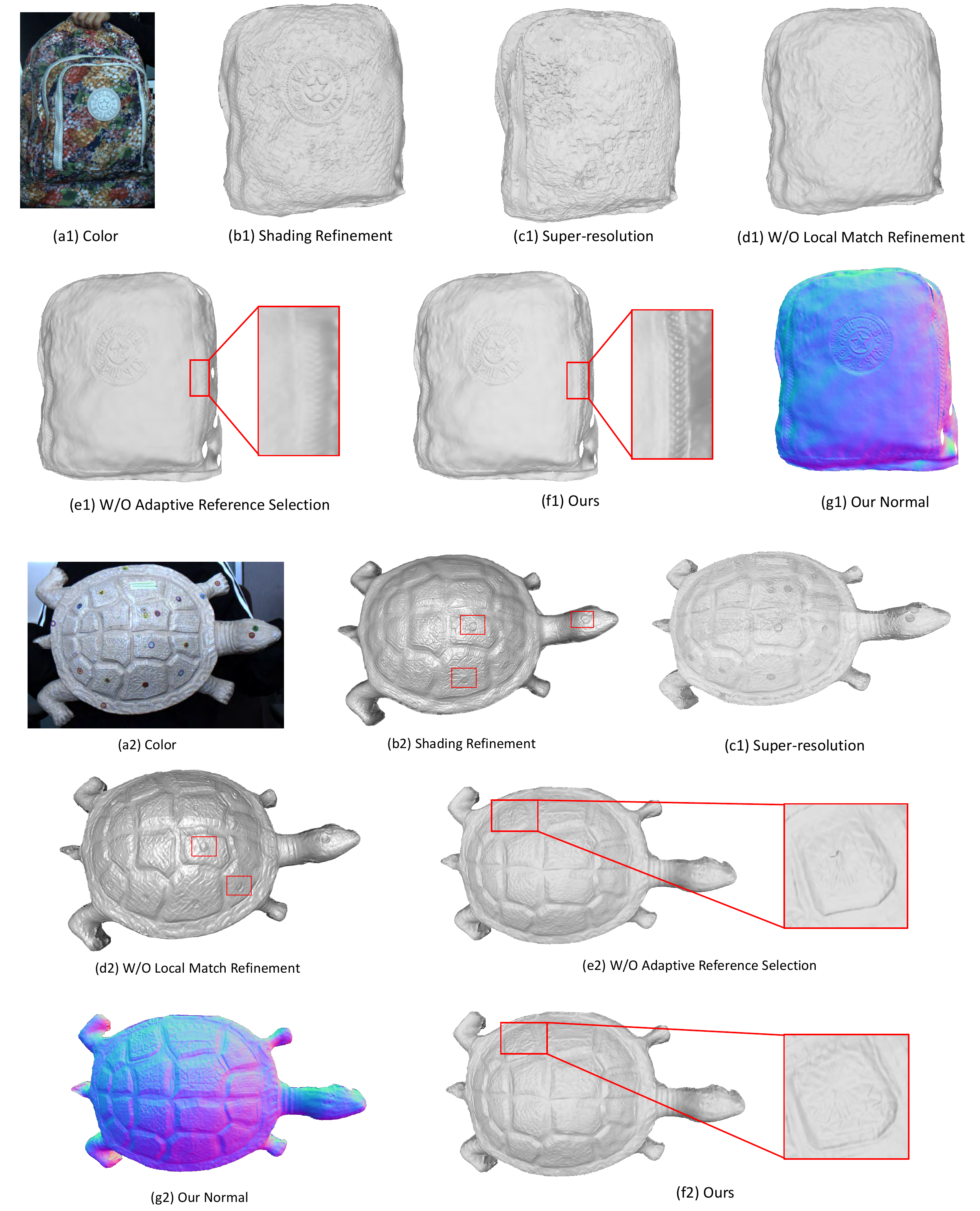}
	\end{center}
	\caption{Comparison results on Backpack and Turtle model. (a1) and (a2) are the reference color images of Backpack and Turtle respectively. The output from shading refinement method~\cite{Or-el15} is shown in (b1) and (b2). The results computed by depth super-resolution~\cite{Haefner18} are displayed in (c1) and (c2). (d1) and (d2) are the meshes acquired using our method but without applying our locally robust matching procedure. (e1) and (e2) are the output meshes of our method but without applying our adaptive reference selection procedure. Finally, (f1) and (f2) are the meshes achieved by our approach with all the components enforced. The normal map is given in (g1) and (g2).}
	\label{Fig:Bagturtle}
\end{figure*}

Fig.~\ref{Fig:Bagturtle} shows the comparison results of a very colorful Backpack and Turtle model and in this figure we demonstrate the effectiveness and importance of our local match refinement step and also the adaptive selection procedure. For the Backpack, it actually experiences non-rigid deformation during the movement and therefore we only consider the front part of the backpack which is mostly rigid. We have marked some colorful patterns on the Turtle surface to make the texture more complex to show the superior performance of our pixel-wise recovery method.

Similar to the Frog and Shoe model, the shading refinement approach~\cite{Or-el15} suffers from the texture-copy problem as shown in (Fig.~\ref{Fig:Bagturtle}(b1) and (b2)). Fig.~\ref{Fig:Bagturtle}(c1) and (c2) displays the super-resolution results~\cite{Haefner18}, for which the surface details have not got recovered clearly. Fig.~\ref{Fig:Bagturtle}(d1) and (d2) shows our results without applying our local match refinement step. The uneven surface in some part is caused by misalignment. We are able to eliminate the artifacts after our locally matching step with real geometric details revealed as shown in Fig.~\ref{Fig:Bagturtle}(f1) and (f2). Fig.~\ref{Fig:Bagturtle}(e1,e2) shows our results without applying the adaptive reference selection step. The results are achieved by setting the image shown in Fig.~\ref{Fig:Bagturtle}(a1) and (a2) as the reference frame. As we can see, for the Backpack model, the surface details on the side of the bag have not got fully recovered without performing the adaptive selection procedure since it hasn't got captured clearly in the selected image as shown in Fig.~\ref{Fig:Bagturtle}(a1). For the Turtle model, as demonstrated in Fig.~\ref{Fig:Bagturtle}(f2) we are able to eliminate small artifacts and get more clear surface details after performing our adaptive selection procedure.

\begin{figure}[!ht]
\begin{center}
   \includegraphics[width=0.825\linewidth]{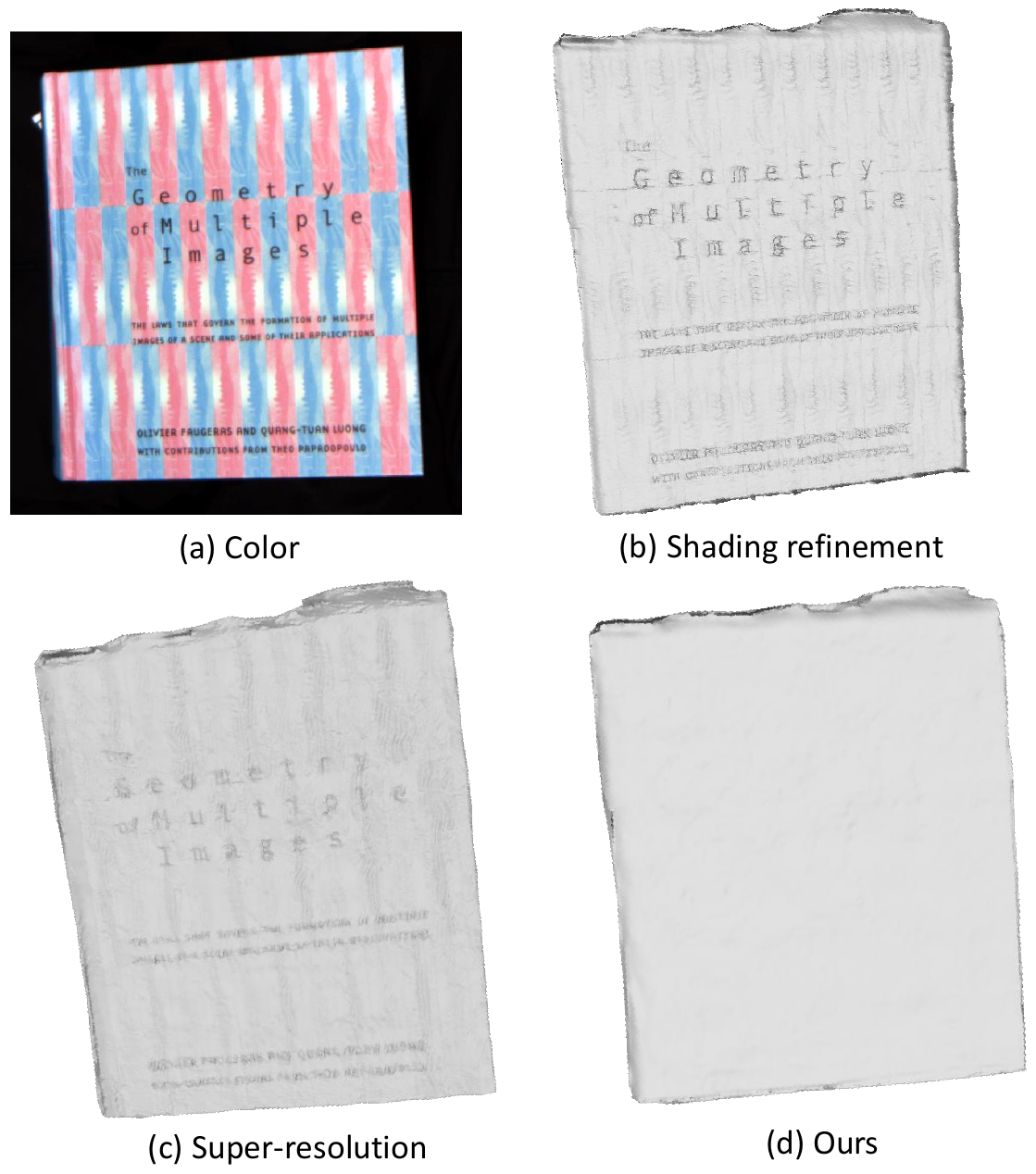}
\end{center}
   \caption{Results on Book model. (a) is the reference color image.  (b) shows the refined mesh with shading refinement method~\cite{Or-el15}. (c) shows the super-resolution results~\cite{Haefner18}. The recovered mesh surface from our method is displayed in (d).}
   \label{Fig:Book}
\end{figure}

To further validate our robustness against texture copy problem, the results of a Book cover with extremely rich textures are demonstrated in Fig.~\ref{Fig:Book}. As displayed in Fig.~\ref{Fig:Book}(d) the textures have been successfully factored out from the image with our approach and the recovered model keeps as a planar surface after the enhancement.
In comparison, the result from shading refinement method (Fig.~\ref{Fig:Book}(b)) and depth super-resolution approach (Fig.~\ref{Fig:Book}(c)) are severely affected by the texture copy effect with lots of fake geometric details appeared.

\begin{figure}[!]
	\begin{center}
		\includegraphics[width=0.90\linewidth]{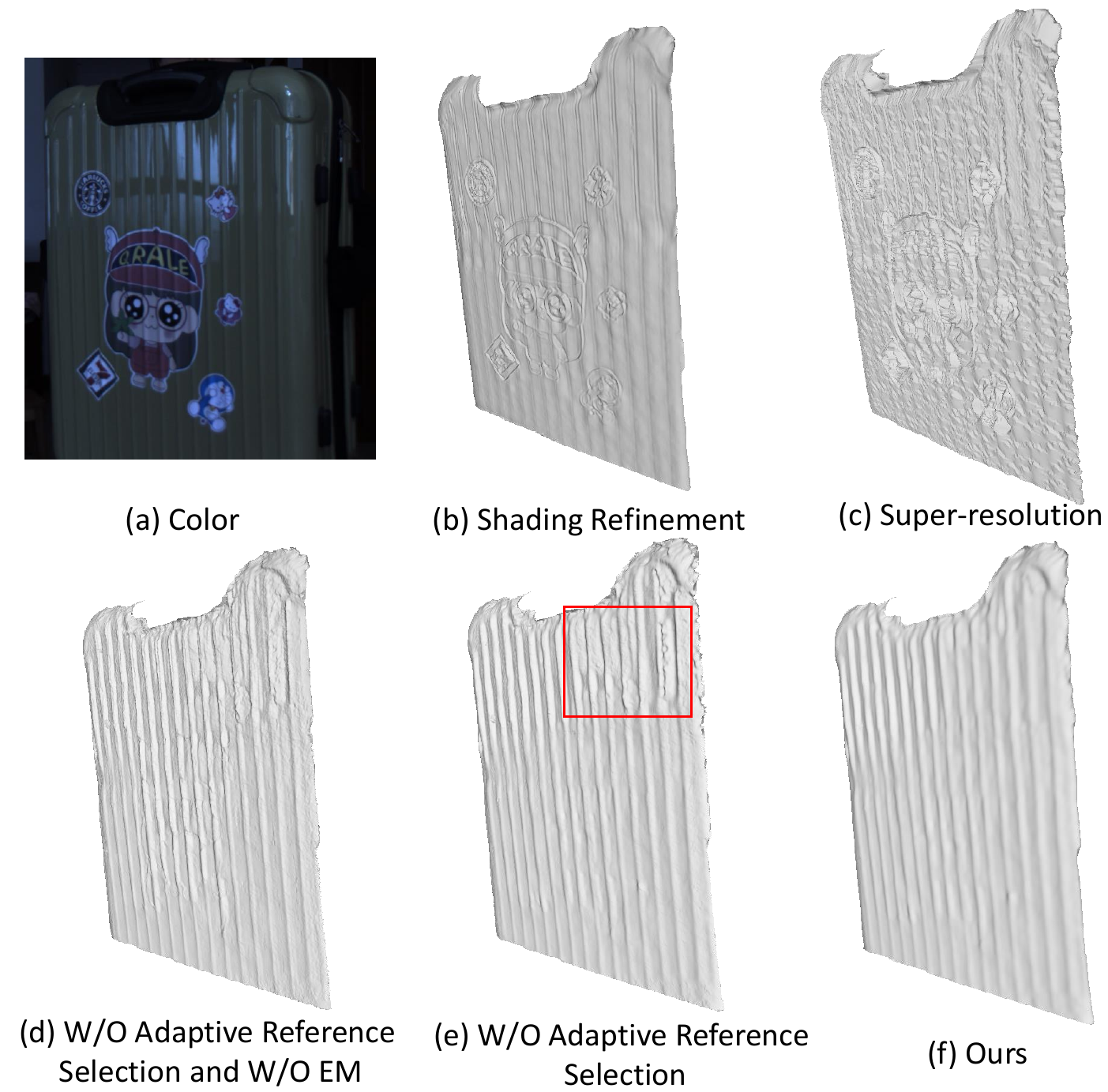}
	\end{center}
	\caption{Comparison results on Suitcase model. (a) is the reference color image. The results computed by shading refinement method~\cite{Or-el15} is displayed in (b). The output from Super-resolution~\cite{Haefner18} result is shown in (c). (d) is the mesh acquired using our method but without applying our adaptive reference selection and EM procedure. (e) is the result after EM procedure but without adaptive selection. Finally, (f) is the mesh achieved by our approach.}
	\label{Fig:adaptiveCmp}
\end{figure}

To better demonstrate the effectiveness of our adaptive reference frame selection and the robustness of our EM based normal recovery approach on handling non-Lambertian reflection such as specularities, in Fig.~\ref{Fig:adaptiveCmp} we show the comparison results of a Suitcase which is not purely Lambertian surface. Fig.~\ref{Fig:adaptiveCmp}(d) displays the result without applying both adaptive reference selection and EM based normal recovery procedure. The surface is quite noisy as affected by the highlight in the sequence. The recovered surface has got better (Fig.~\ref{Fig:adaptiveCmp}(e)) after our EM optimization approach which has filtered out some outliers during the normal computation. However, the artifacts have not got fully handled (marked in a red box in Fig.~\ref{Fig:adaptiveCmp}(e)) if we just take the image in Fig.~\ref{Fig:adaptiveCmp}(a) as the reference image instead of performing the adaptive reference selection method. Fig.~\ref{Fig:adaptiveCmp}(f) shows our final result with all the components enforced, which is more pleasant and robust to imperfect Lambertian reflectance. 

\subsubsection{Intrinsic Image Decomposition}
\begin{figure}[!h]
\begin{center}
   \includegraphics[width=0.9999\linewidth]{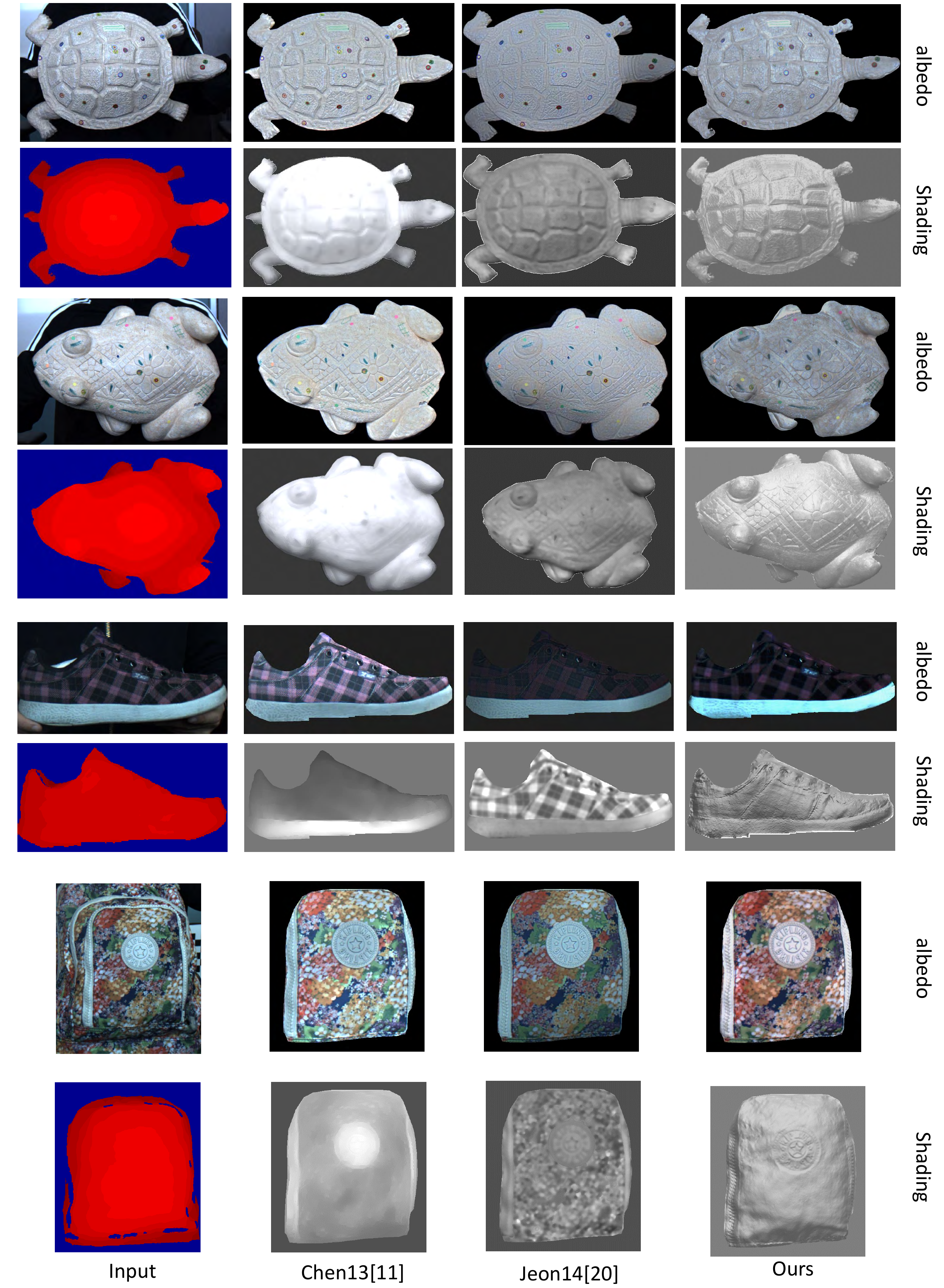}
\end{center}
   \caption{Comparison results on albedo recovery or intrinsic decomposition of the Turtle, Frog, Shoe and Backpack model. The first column is the input color image with its corresponding depth map. The second column shows the result of Chen~\cite{Chen13}. The third column is the decomposed albedo and shading images from method in~\cite{Jeon14}. Finally, the last column demonstrates the result achieved by our method.}
   \label{Fig:intrinsic}
\end{figure}

In order to show the performance of our method in albedo recovery, we have also made some comparisons with two state-of-the-art intrinsic image decomposition approaches~\cite{Chen13,Jeon14} as displayed in Fig.~\ref{Fig:intrinsic}. For these two compared methods, they take the RGB-D images of the reference frame as input, as displayed in the first column. The second column shows the result from Chen~\cite{Chen13}, for which the shading image is over smoothed with the geometry details decomposed into albedo map incorrectly. The method from Jeon~\cite{Jeon14} has better results on recovered shading images for the Turtle and Frog models as displayed in the third column. However, some textures still stay at the shading image especially for the Shoe and Backpack. In comparison, with our pixel-wise albedo computation method, we are able to recover a much sharper albedo map and the ``texture copy'' effect in the geometry is barely noticeable.

%% file: conclusion.tex
\vspace{-5pt}
\section{Conclusion}

In this paper, we present a novel approach to recover surface details and its albedo map from an RGB-D video sequence. The object is experiencing casual motion from which the induced illumination variation provides us the cue to recover the surface normal and its albedo as well. We have proposed a labeling approach to select reference images adaptively along the sequence. Then a robust lighting insensitive local match strategy is proposed to establish correct correspondences from reference frame to other frames. Then, the environmental lighting is estimated by exploiting the whole sequence to get rid of the effect of varying textures. Finally, the surface normal and its albedo is calculated robustly with our EM framework. We have validated our method on both synthetic and real datasets and compared with some state-of-the-art surface refinement and intrinsic decomposition methods. As demonstrated in the experiments, we have achieved good performance on both surface details recovery and intrinsic decomposition.

\textbf{Limitations.} Our approach works well on convex objects that come with a whole piece of surface and it is challenging to deal with objects with great concavities and discontinuities since it will be difficult to compute a continuous warping field to get precise alignment due to self-occlusion. More importantly the concave part will probably get occluded during the capture as we rotate the object. In this case, we will not be able to collect enough evidence of the surface under different lighting condition, causing surface normals to fail to converge to its optimal value. Also, our method fails on objects that are too small since the fusion step will fail in the first place. Another limitation is that we assume that the environment lighting do not have sudden changes as we capture the image sequence. The formulation will become invalid when the lighting changes which means the changes of the pixel intensity are not just caused by the object rotation. We will have to take both the change of environmental lighting and object motion into account. As a future work, we could take advantages of the changing lighting conditions in our formulations since more lighting variations reveal more valuable information about the surface.
Finally, right now we are mainly focusing on depth enhancement, while we would like to implement all these procedures in full 3D space and recover a complete model.
 
